\documentclass{llncs}

\usepackage[usesms]{stex}
\usepackage{stex-highlighting}
\usepackage{theorems}
\usepackage{amsmath}

\usepackage[outputdir=., cachedir=., frozencache]{minted}
\setminted[text]{breaklines, breakafter={\{]}, breakbefore=\}, escapeinside=!!, breakbeforesymbolpre=, breakaftersymbolpre=}
\usepackage{url}
\usepackage{wrapfig}
\usepackage{multicol}
\usepackage[dvipsnames]{xcolor}
\usepackage{graphicx}
\usepackage{cleveref}
\usepackage{setspace}
\usepackage{tabularx}
\usepackage[skip=3pt]{caption}
\usepackage{forest}
\usepackage{amsfonts}
\usepackage{vwcol}
\usepackage{mdframed}
\usepackage{bbding}
\usepackage{orcidlink}

\newenvironment{ttmath}
 {\everymath{\ttmathgroup}\everydisplay{\ttmathgroup}}
 {}

\forestset{
  my tier/.style={
    tier/.wrap pgfmath arg={level##1}{level()},
  },
}
\forestset{
    indentation/.style={for tree={
        grow'=0,
        child anchor=west,
        parent anchor=south,
        anchor=west,
        calign=first,
        s sep+=-6pt,
        inner sep=2.3pt,
        edge path={
        \noexpand\path [draw, \forestoption{edge}]
        (!u.south west) +(7.5pt,0) |- (.child anchor)\forestoption{edge label};
        \forestoption{edge=no edge}
        },
        before typesetting nodes={
        if n=1{
            insert before={[, phantom, my tier]},
        }{},
        },
        my tier,
        fit=band,
        before computing xy={
        l=12pt,
        },
    }
    }
}

\newmdenv[linecolor=black]{mdframe}

\title{Towards Semantic Markup of Mathematical Documents via User Interaction}
\institute{Heriot-Watt University, Edinburgh, United Kingdom\linebreak
\email{\{lv21, F.D.Kamareddine\}@hw.ac.uk}}

\author{Luka Vrečar\Envelope\inst{1}\orcidlink{0009-0000-0927-6159} \and Joe Wells\inst{1} \and Fairouz Kamareddine\inst{1}\orcidlink{0000-0002-6141-2709}}

\begin{document}
\maketitle
$\mathtt{\xdef\ttmathgroup{\fam\the\fam\relax}}$
\usemodule[smglom/trigonometry]{mod?sine-series}

\usemodule[LVrecar/lcalc]{mod?variables}
\usemodule[LVrecar/lcalc]{mod?setoflambdaexpressions}
\usemodule[LVrecar/lcalc]{mod?lcalc}
\usemodule[LVrecar/lcalc]{mod?meta}

\usemodule[smglom/mv]{mod?equal}
\usemodule[smglom/mv]{mod?mv}

\usemodule[smglom/linear-algebra]{mod?matrix}

\usemodule[smglom/sets]{mod?set}
\usemodule[smglom/grammar]{mod?grammar}

\usemodule[sTeX/MathTutorial]{mod?Nat}
\begin{smodule}{gui-report}
\notation{gseq}[space]{\argsep{#1}{\;}}
\setnotation{gseq}{space}
\symdef{galt}[name=grammar-alternatives, args=a]{\argsep{#1}{\;|\;}}
\def\lambdaterm{$\lambda$-term}
\def\lambdaterms{$\lambda$-terms}
\def\parglare{{\footnotesize\texttt{parglare}}}
\def\latexwalker{{\footnotesize\texttt{latexwalker}}}
\setlength{\intextsep}{0pt}

\begin{abstract}
    Mathematical documents written in \LaTeX\ often contain ambiguities.
    We can resolve some of them via semantic markup using, e.g., \sTeX, which also has other potential benefits, such as interoperability with computer algebra systems, proof systems, and increased accessibility.
    However, semantic markup is more involved than ``regular'' typesetting and presents a challenge for authors of mathematical documents.
    We aim to smooth out the transition from plain \LaTeX\ to semantic markup by developing semi-automatic tools for authors.
    In this paper we present an approach to semantic markup of formulas by (semi-)automatically generating grammars from existing \sTeX\ macro definitions and parsing mathematical formulas with them.
    We also present a GUI-based tool for the disambiguation of parse results and showcase its functionality and potential using a grammar for parsing untyped \lambdaterms.
\end{abstract}

\section{Introduction}
Formulas in mathematical documents written in \LaTeX\ are often ambiguous.
We can disambiguate them using \sTeX \cite{kohlhaseSTEX3PackageCollection2023,kohlhaseSTEXdescription}, a package for \textit{semantic markup} of mathematical documents using \textit{semantic macros}.
This allows the explicit encoding of the semantics of formulas without changing their typeset appearance.
\begin{mdframe}
    \begin{sexample}[title={The ambiguity of $\cart{P,Q}$}] \label{example: ambiguity of PxQ}
        The formula $\cart{P, Q}$ is usually typeset using the ambiguous typesetting {\footnotesize\verb|P \times Q|} (where the meaning of {\footnotesize\verb|\times|} depends on the type of $P$ and $Q$).
        Instead one can use the semantic macro {\footnotesize\verb|\cart{P,Q}|} to denote the \sr{cart}{Cartesian product} of two sets, or {\footnotesize\verb|\matrixtimes[x]{P,Q}|} to denote the multiplication of two matrices.
    \end{sexample}
\end{mdframe}
We used \sTeX\ to typeset the example above.
Note that this also generates hyperlinks to online resources\footnote{Coloured text denotes such hyperlinks (opens a new browser tab or jumps to the relevant definition in the PDF). Blue is used for highlighting notational components of macros, and teal is used for textual references to macros. Magenta colouring is used in the online definitions to highlight the defining occurrence of a symbol.}.
Semantic markup can facilitate clearer communication with readers, e.g., students or people with disabilities.
It can also support connections between human-readable documents and proof systems (for formal verification), computer algebra systems (for computation), etc. 
Semantically marked up documents can be dynamic\footnote{Not necessarily in PDF, but other formats like HTML support dynamic documents with ease.} and adapt to user preferences and interaction.
\begin{mdframe} 
    \begin{sexample}[title={Semantic markup for encoding the structure of a formula}] \label{example: semantic markup for formula structure}
        Consider the formula $\sin x/y-z$ \cite{kohlhaseCoRepresentingStructureMeaning}, typeset as {\footnotesize\verb|\sin x/y-z|}.
        This can be interpreted as $\sin(\frac{x}{y})-z$, $\sin(\frac{x}{y-z})$, or even $\frac{\sin x}{y-z}$.
        A reader can disambiguate this with sufficient background knowledge and context, e.g. some of the readings can be nonsensical if $y-z=0$. Additionally, implicit parentheses can be ``filled in'' based on prior reader knowledge (in the case of $\sin$, we usually write it without parentheses when the argument is a monomial).
        \notation{sine}[nb]{\comp{\sin}\;#1}
        \notation{realminus}[nb]{\argsep{#1}{\mathbin{\comp-}}}
        \notation{realdivide}[nb]{#1\comp{/}#2}
        
        We could use, e.g., {\footnotesize\mintinline[breaklines=true, breakbefore={\{}, breakafter={,}]{text}|\realminus{\sine{\realdivide{x}{y}},z}|}, to reflect the intended structure using \sTeX, which would be typeset as $\realminus[nb]{\sine[nb]{\realdivide[nb]{x}{y}},z}$ and can be read equivalently to $\realminus{\sine{\realdivide[frac]{x}{y}}, z}$.
        This precisely encodes the intended structure of the formula while also offering all of the advantages of \sTeX, such as hyperlinks to online resources.
        In an active document, we can then add ways to show the structure of the formula (e.g., by rendering it as a tree), or automatically insert brackets for disambiguation\footnotemark.
        
    \end{sexample}
\end{mdframe}
\footnotetext{Automatic bracketing is already a feature in \sTeX, see \cite{kohlhaseSTEX3PackageCollection2023}, section 7.4.1.}
Semantic markup clearly has advantages over plain \LaTeX, but producing documents via \sTeX\ is more involved.
Our goal is to facilitate a smooth transition to \sTeX\ for new users, by devising ways to (semi-)automatically add semantic markup to \LaTeX\ documents.
We will refer to this as \sTeX-ification, or \sTeX-ifying a document.

The initial focus of our work are documents which do not contain semantic annotations.
We sampled a number of papers from a number of areas of mathematics and computer science and found a complete absence of macros that record the meaning or structure of formulas being used.
Our conclusion was that a significant number of documents does not not contain semantic annotations, so it made sense to focus on those for now, and explore handling documents with existing (author-defined) semantic annotations in future work.

Authors of documents without semantic annotations (especially those involved in teaching) likely do not have experience with them, but could benefit from the features built around \sTeX\ such as the hyperlinks to definitions, definitions on hover\footnote{In HTML documents, which can be generated from a \LaTeX\ source using RusTeX \cite{mullerSlatexRusTeX2024}} and ALeA \cite{bergesLearningSupportSystems2023}.
By offloading the actual semantic annotating (i.e., rewriting of formulas to use semantic macros) we aim to provide the benefits of semantic annotations, without authors having to learn \sTeX\ beyond the very basics (compiling documents with \sTeX, and defining new semantic macros/notations).
We also aim to provide a way of annotating a document for authors who are not familiar with fast text-editing techniques like Emacs macros.

\begin{table}
\begin{mdframe}
    \caption{The \sTeX-ification process} \label{table: stexification process}
    \begin{enumerate}
        \item Generating all the prerequisites \begin{enumerate}
            \item Manually identify all the semantic macros required, and define any new macros that might not be available in existing collections.
            \item Create a grammar for parsing all the formulas. This can be done manually, or (semi-)automatically (as described in \Cref{section: grammar generation}).
        \end{enumerate}
        \item Actually \sTeX-ifying the document \begin{enumerate}
            \item Parse each formula in the \LaTeX\ source with the grammar from 1b.
            \item Disambiguate any parses by prompting the user to select the correct parse tree via a graphical interface we present in \Cref{section: gui},
            \item Convert all the parse trees into \sTeX\ macros.
            \item Create a copy of the original document, with each formula replaced by its \sTeX\ counterpart.
        \end{enumerate}
    \end{enumerate}
\end{mdframe}
\end{table}

This paper explores (semi-)automatic semantic annotating and the problems that must be solved to make our approach feasible for large documents (e.g., lecture notes or papers).
We propose a method for automatically generating grammars from \sTeX\ macro definitions, parsing formulas with them, and using a graphical user interface (GUI) for disambiguation.
In \Cref{section: background}, we describe \sTeX\ in more detail, and also introduce the untyped \lcalc, which we are using as a testing ground for our methods.

We envision \sTeX-ification as a two-stage process (see \Cref{table: stexification process}).
\Cref{section: groundwork} describes the first stage, which involves identifying the macros we need to \sTeX-ify a given document (stage 1a), and creating a grammar for parsing the formulas within it (stage 1b).
The grammar can be manually specified or automatically generated (see \Cref{section: grammar generation}) based on the users' selection of relevant \sTeX\ modules\footnote{Semantic macro definitions in \sTeX\ are organized into \textit{archives} (for areas of mathematics, e.g., calculus) and \textit{modules} (for defining macros for a specific mathematical object, e.g., the integral).}.
We also present the \parglare\ parsing library \cite{DEJANOVIC2022102734} we are using.
\Cref{section: gui} describes the second stage, which involves parsing individual formulas (stage 2a) and using our GUI-based tool for disambiguating parses (stage 2b).
Once all the formulas are disambiguated, the parse trees are converted into \sTeX\ macros and a copy of the document is created, where formulas are replaced with their \sTeX-ified counterparts (stage 2d).
We demonstrate this with an example grammar (see \Cref{example: grammar for lambda terms}).
\Cref{section: grammar generation} showcases the current state of our automatic grammar generation, which was developed to make stage 1 of \sTeX-ification easier.
\Cref{section: conclusion} summarizes the work and lists some future improvements.

\section{Background} \label{section: background}
This section outlines the background needed to contextualize the research.
We assume familiarity with context-free grammars (CFGs) \cite{ahoCompilersPrinciplesTechniques2006} and GLR parsers \cite{tomitaGeneralizedLRParsing1991}.
This section gives a brief introduction to \sTeX\ and the (untyped) \lcalc\footnote{We will use the phrases ``untyped \lcalc'' and ``\lcalc'' interchangeably from this point on.}.

\subsection{\sTeX}

The \sTeX\ package \cite{kohlhaseSTEX3PackageCollection2023} was developed by the KWARC research group.
Its main feature is a systematic way to generate \textit{semantic macros}, which allow users to semantically annotate their mathematical documents.
This allows for the explicit recording of, e.g., structure in a mathematical expression.
The group also started the development of the Semantic Multilingual Glossary of Mathematics (SMGloM) \cite{kwarcSMGloMGitLab}, which contains semantic macro definitions for hundreds of concepts from various areas of mathematics.
The SMGloM also provides natural language definitions of those concepts in English, and sometimes other languages (German being the second most common).
The package is well documented in \cite{kohlhaseSTEX3PackageCollection2023}.
Nonetheless we will reference some features which we use in this paper.

By semantically marking up terms using \sTeX\ we can also use the systems surrounding it, like the automatically generated hyperlinks to online definitions.

\subsubsection{Previous Attempts at Automatic \sTeX-ification}
A notable previous attempt at automatic \sTeX-ification was by Müller and Kalyszyk \cite{mullerDISAMBIGUATINGSYMBOLICEXPRESSIONS2021}, who trained a machine learning (ML) model on a corpus of \LaTeX\ fragments and their  semantically annotated counterparts produced using \sTeX.
They achieved some success with GPT-2, producing correct \sTeX-ifications around 50\% of the time.

Instead of probability-based predictions, we present an approach that relies on user input for correct \sTeX-ification.
This also removes the need for training an ML model, while being sufficiently flexible to handle arbitrary semantic macro definitions without large modifications to the system we developed. 

\subsection{The (Untyped) \lcalc}
To demonstrate the methods we propose in this paper, we need to choose an area of mathematics for which we want to \sTeX-ify documents.
The overall problem of semantically annotating mathematical formulas is present in all areas of mathematics, but the focus of this paper will be the (untyped) \lcalc.
We do this because we are the teaching team of the Foundations course on the \lcalc, which is a subject of our experimentation in \sTeX-ifying its entire \LaTeX\ library.
We wish to extend tests to our students across all our campuses (around 250).
That being said, here is a short recap of the \lcalc:
\def\bvar#1{\textbf{#1}}
\begin{mdframe}
\begin{sdefinition}[title=Variables, for={var, varSet}] \label{def: variables}
    Let $\varSet$ be the set of variables, defined as $\eq{\varSet, \set{\bvar{v}, \bvar{v'}, \bvar{v''}, \ldots}}$.
    We will denote the \textit{meta-variables} that range over $\varSet$ with lowercase letters (e.g., $\var{x}$, $\var{y}$, $\var{z}$), that can have apostrophes or subscripted index attached (e.g., $\var{x'}$, $\var{y_{1}}$, $\var{z_{2}''}$).
\end{sdefinition}
\end{mdframe}
\begin{mdframe}
\begin{sdefinition}[title={The set of \lambdaterms}, for={setOfLambdas, abs, app}]
    Let $\setOfLambdas$ be the set of all \lambdaterms.
    We will denote the \textit{meta-terms} that range over $\setOfLambdas$ with uppercase letters (e.g., $A$, $B$, $C$), that can have apostrophes or subscripted index attached (e.g., $A'$, $B_{1}$, $C_{2}''$).
    We define $\setOfLambdas$ inductively as follows:
    \begin{itemize}
        \item If $\inset{\var{x}}{\varSet}$, then $\var{x}$ is in $\setOfLambdas$.
        \item If $\inset{A, B}{\setOfLambdas}$, then the \sr{app}{\textit{application of $A$ to $B$}}, denoted by $(\app{A}{B})$, is in $\setOfLambdas$.
        \item If $\inset{\var{x}}{\varSet}$ and $\inset{A}{\setOfLambdas}$, then the \sr{abs}{\textit{abstraction in $A$ over $\var{x}$}}, denoted by $(\abs{\var{x}}{A})$, is in $\setOfLambdas$.
    \end{itemize}
\end{sdefinition}
\end{mdframe}
\begin{mdframe}
\begin{sdefinition}[title={Notational conventions of the \lcalc}] \label{def: notational conventions of the lcalc}
    We employ some notational conventions when writing out \lambdaterms.
    We follow the conventions from our Foundations course notes:
    \begin{itemize}
        \item We can remove the outermost parentheses in a term: we can write $\app{A}{B}$ instead of $(\app{A}{B})$.
        \item Application is left-associative: we can write $\app{(\app{A}{B})}{C}$ as $\app{\app{A}{B}}{C}$.
        \item The scope of an abstraction extends as far to the right as possible: $\abs{x}{\app{x}{y}}$ is equivalent to $\abs{x}{(\app{x}{y})}$, NOT $\app{\abs{x}{x}}{y}$.
        \item Multiple consecutive abstractions can be ``compressed'': we can write $\abs{x}{(\abs{y}{(\abs{z}{A})})}$ as $\abs{x, y, z}{A}$. \label{ref: compressing abstractions}
    \end{itemize}
\end{sdefinition}
\end{mdframe}

\section{Laying the Groundwork} \label{section: groundwork}
\def\demofile{{\footnotesize\texttt{demo-file.tex}}}
In this section, we present the first stage of our approach to \sTeX-ification (see \Cref{table: stexification process}).
Throughout this section, we will assume that we wish to \sTeX-ify a document, whose formulas contain only untyped \lambdaterms\ (let us call it \demofile).
Its contents are shown in \Cref{example: demofile contents}.
In \Cref{section: parglare} we will also present the parsing library we are using.

\subsection{Identifying the Required \sTeX\ Macros} \label{section: stex macros for the lcalc}
The goal of this section is to define some macros and a grammar in order to \sTeX-ify \demofile.
\Cref{table: semantic macro definitions} lays outsome semantic macros we defined to semantically annotate \lambdaterms.

\setlength{\intextsep}{0pt}
\begin{wraptable}{r}{\textwidth}
\begin{mdframe}
    \caption{The macros we define for semantically annotating \lambdaterms.}
    \label{table: semantic macro definitions}
    \begin{itemize}
        \item {\footnotesize\mintinline[breaklines=true]{text}|\var{#1}|} - marks up its argument as a variable.
        \item {\footnotesize\mintinline[breaklines=true]{text}|\abs{#1}{#2}|} - the argument 1 is a \textit{flexary argument}\footnotemark{} representing a sequence of variables, and argument 2 is the body of the abstraction.
        \item {\footnotesize\mintinline[breaklines=true]{text}|\app{#1}{#2}|} - both arguments are terms and this denotes the application of the first to the second.
    \end{itemize}
\end{mdframe}
\end{wraptable}
\def\tablewidth{0.68\textwidth}

\begin{wraptable}[9]{r}{\tablewidth}
\begin{mdframe}
    \caption{A side by side comparison of \LaTeX\ source code with \sTeX\ macros and the typeset results.}
    \begin{tabular}{l|l}
        \hline
        Source code & Typeset result \\ \hline
        {\footnotesize\mintinline[breaklines=true]{text}|\var{x}|} & $\var{x}$ \\ \hline
        {\footnotesize\mintinline[breaklines=true, breakbefore={\{}, breakafter={,}]{text}|\abs{\var{x},\var{y},\var{z}}{A}|} & $\abs{\var{x}, \var{y}, \var{z}}{A}$\\ \hline
        {\footnotesize\mintinline[breaklines=true]{text}|\app{A}{B}|} & $\app{A}{B}$ \\ \hline
    \end{tabular}
    \label{table: stex examples}
\end{mdframe}
\end{wraptable}
\footnotetext{These are arguments of variable length, supplied as comma-separated values.
For more details, we refer the reader to the \sTeX\ documentation \cite{kohlhaseSTEX3PackageCollection2023}, section 3.1.3.}
\Cref{table: stex examples} shows examples of source code using the macros, and the typeset results.
Our aim is to use our macros and tools to \sTeX-ify the material for our course at Heriot-Watt.
Having defined the macros we need to \sTeX-ify \demofile, we can now create a grammar to parse the formulas within the document.

\subsection{Parsing Formulas with \parglare} \label{section: parglare}
For our parsing, we are using \parglare\ \cite{DEJANOVIC2022102734}, a GLR parser written in Python.
GLR parsers are suitable for our parsing needs as they are exhaustive and support ambiguous grammars.
Parsing terms with \parglare\ produces parse trees, which we can transform into abstract syntax trees (ASTs) using \textit{parse actions}.
\begin{wrapfigure}[15]{r}{0.6\textwidth}
\begin{mdframe}
    \begin{footnotesize}
        \begin{vwcol}[widths={0.6, 0.4}, rule=0pt]
            \begin{forest}
                for tree={l=15pt}
                [\texttt{lexp}
                    [\texttt{parexp}
                        [$($]
                        [\texttt{lexp}
                            [\texttt{abs}
                                [\texttt{lam}
                                    [\textbackslash lambda]
                                ]
                                [\texttt{varlist}
                                    [\texttt{var}
                                        [$x$]
                                    ]
                                ]
                                [\texttt{dot}
                                    [.]
                                ]
                                [\texttt{var}
                                    [$x$]
                                ]
                            ]
                        ]
                        [$)$]
                    ]
                ]
            \end{forest}
            
            \begin{forest}
                for tree={l=15pt}
                [dobrackets
                    [abs
                        [varlist
                            [var
                                [$x$]
                            ]
                        ]
                        [var
                            [$x$]
                        ]
                    ]
                ]
            \end{forest}
        \\
        \end{vwcol}
    \end{footnotesize}
    \captionsetup{skip=2\baselineskip}
    \caption{The parse tree and AST for $\protect(\abs{\var{x}}{\var{x}})$} \label{fig: parse tree and AST}
\end{mdframe}
\end{wrapfigure}
Parse actions are a \parglare\ feature which allows for transforming of parse trees into other forms, either by using the built-in functions provided by \parglare, or defining custom ones.
We use custom parse actions to build ASTs that map easily onto the \sTeX\ macros.
They are built so that the names of nodes match the names of semantic macros identified in \Cref{section: groundwork}, or native \sTeX\ macros like \texttt{dobrackets}, which wraps its argument in parentheses.

Once parsing is complete, we can transform the ASTs into a series of macro calls and their arguments to create semantic annotations.
This is relatively straightforward, as the nested structure of semantic macros is similar to the ASTs produced during parsing.
\Cref{fig: parse tree and AST} shows the parse tree created by parsing $(\abs{\var{x}}{\var{x}})$ using the grammar in \Cref{example: grammar for lambda terms}, and the AST produced by running our parse actions over it.
Text in \texttt{\footnotesize typewriter} font denotes nonterminals.

\begin{mdframe}
\begin{sexample}[title={The grammar used for parsing \lambdaterms}]\label{example: grammar for lambda terms}
    \begin{footnotesize}
        \setlength{\jot}{2pt}
        \begin{ttmath}
            \begin{align*}
                & \grule{lexp}{\galt{app, var, abs, parexp}} & \text{\texttt{lexp} matches any \sr{lterm}{$\lambda$-term}} \\
                & \grule{app}{\gseq{lexp, lexp}} & \text{\texttt{app} matches \sns{app}} \\
                & \grule{abs}{\gseq{lam, varlist, dot, lexp}} & \text{\texttt{abs} matches \sns{abs}} \\
                & \grule{lam}{\galt{``\lambda", ``\textbackslash lambda"}} & \text{\texttt{lam} matches either } \lambda \text{ or \texttt{\textbackslash lambda}} \\
                & \grule{varlist}{\galt{\gseq{var, varlist}, var}} & \text{\texttt{varlist} matches a list of variables} \\
                & \grule{parexp}{\gseq{``(", lexp ,``)"}} & \text{\texttt{parexp} matches a \texttt{lexp} in parentheses} \\
                & \grule{var}{[a-z]+?} & \text{\texttt{var} is a regex matching single letters} \\
                & \grule{dot}{``."} & \text{\texttt{dot} matches a single ``.'' \footnotemark{}}
            \end{align*}
        \end{ttmath}
    \end{footnotesize}
\end{sexample}
\end{mdframe}
\footnotetext{This is due to a limitation of the parsing library used, see \Cref{section: limitations of parglare}.}

The grammar we define is ambiguous, as it has no notion of associativity of application, or how far the scope of an abstraction extends.
That means that it parses terms like $\app{\app{\var{x}}{\var{y}}}{\var{z}}$, and $\abs{\var{x}}{\app{\var{y}}{\var{x}}}$ ambiguously\footnote{The aim of this paper isn't to create an unambiguous grammar for parsing \lambdaterms.
Those already exist in the literature, but an unambiguous grammar would not showcase all the features of our other work, such as the GUI (see \Cref{section: gui}).}.

The grammar was designed by hand, which can be a lengthy process and would not scale to more areas of mathematics easily.
We propose a solution in the form of (automatic) grammar generation, described in \Cref{section: grammar generation}.

\subsubsection{Recognizers}
Another powerful feature of \parglare\ are its \textit{recognizers}.
A recognizer is a Python function the user can supply to the parser for tokenizing portions of the input.
For example, matching natural numbers without leading 0s would require a complex regular expression or several grammar rules.
It might be more convenient to write a recognizer for such a task.
We can improve the grammar we defined in \Cref{example: grammar for lambda terms}, by replacing the rule ${\footnotesize\grule{\texttt{var}}{\texttt{[a-z]+?}}}$ with a recognizer for variables with apostrophes or subscripted indexes (to better match the definition of variables in \Cref{def: variables}).
Implementing this with CFG rules is difficult due to \TeX's peculiarities (e.g., optional braces around arguments), but a recognizer can do it with a Python function. This allows us to use libraries for parsing \LaTeX, (e.g., \texttt{\footnotesize pylatexenc} \cite{faistWelcomePylatexencDocumentation2023}), that make the task more manageable.

\subsubsection{Limitations of \parglare} \label{section: limitations of parglare}
The main limitation of \parglare\ that we encountered is its inability to handle cycles in grammars (see \Cref{section: current state and issues} for more detail).
A different, rather small drawback is that a period cannot appear at the start of a string literal in the grammar.
This can be easily resolved by adding a production of the form $\grule{\texttt{dot}}{\texttt{"."}}$ to each grammar (see \Cref{example: grammar for lambda terms}).

\section{A Disambiguation GUI for Parsing during \sTeX-ification} \label{section: gui}
This section describes the second stage of our approach, the disambiguation GUI and its functionality.
It was developed using the Python bindings for GTK \cite{thegtkteamGTKProjectFree2023}.
By developing the GUI in Python, it easily integrates with \parglare.

Once we have identified the macros and created a \parglare\ grammar, we can open our \texttt{.tex} source file within the GUI.
All the formulas within the file are extracted and parsed using the grammar. Any ambiguous parses will show up in the GUI's \textit{formula view} (section 3 in \Cref{fig: main gui with sections}).
The user can then go through all the formulas and select the correct parses (see \Cref{section: gui overview}).
Once all the formulas have been disambiguated in this way, a copy of the original \texttt{.tex} file is created, with all the formulas replaced by their \sTeX\ equivalents.

We have successfully used the GUI to semantically annotate some slides for the Foundations course.
It did not fully \sTeX-ify every formula (the grammar did not cover all of them), but it sped up the process significantly\footnote{Based on the author's subjective experience with manually \sTeX-ifying mathematical documents in the past.}.
\begin{mdframe}
\begin{sexample}[title={The contents of \demofile}] \label{example: demofile contents}\
\begin{footnotesize}
\begin{minted}[breaklines=true]{text}
\documentclass{article}
\begin{document}
Here we have some terms that are ambiguous (according to the grammar we defined).
$\lambda xyz.xy$
Here, the parser cannot decide whether it is an application of $\lambda xy.x$ to $y$ or an abstraction with an application $xy$ inside the body.
$xyzw$ trips up the parser because it is not aware of the left associativity of application.
The issue with $(\lambda xy.xy)$ is that the parser could read this as an application of $\lambda xy.x$ to $y$ wrapped in parentheses, rather than an abstraction (with an application $xy$ inside the body) that is wrapped in parentheses.
\end{document}
\end{minted}
\end{footnotesize}
\end{sexample}
\end{mdframe}
\subsection{Overview} \label{section: gui overview}
The main window of the GUI has 8 sections. \Cref{fig: main gui with sections} shows the GUI with no file opened, and \Cref{fig: the gui with an open file} shows the GUI after \demofile\ has been opened.
The sections of the GUI, as shown in \Cref{fig: main gui with sections}, are: (1) the toolbar, (2) a dropdown for selecting the parsed formula, (3) the \textit{formula view}, showing all the ambiguous formulas as Unicode strings\footnote{This is done to improve readability, as it replaces all \LaTeX\ macros with something more closely resembling their typeset results}, (4) the name of the document being processed, (5) a button to confirm the user's choices, (6) a label displaying the currently selected formula, (7) a grid layout for displaying ASTs, and (8) a ``quit'' button.
\setlength{\intextsep}{0pt}
\begin{figure}[h!]
    \begin{mdframe}
    \begin{tikzpicture}
        \node[anchor=south west,inner sep=0] (image) at (0,0) {\includegraphics[width=\linewidth]{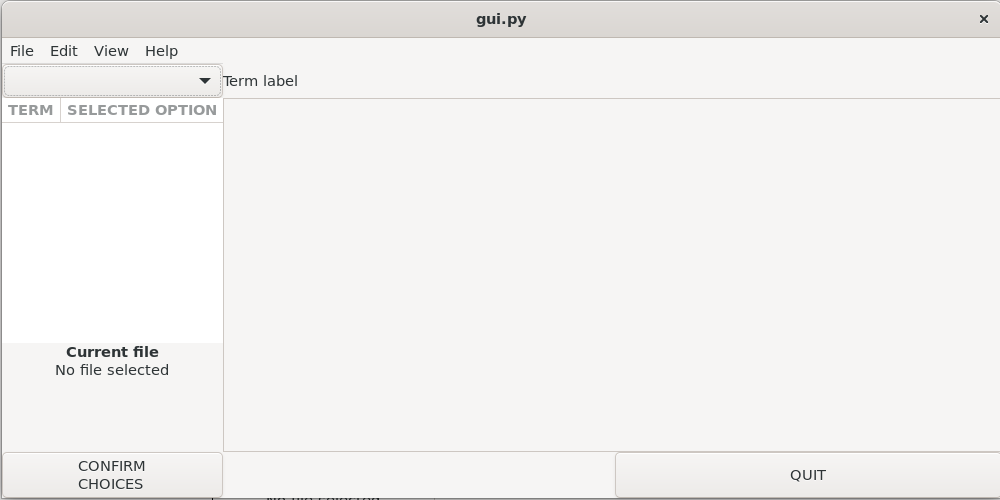}};
        \begin{scope}[x={(image.south east)},y={(image.north west)}]
            \draw[ultra thick,rounded corners] (0.001, 0.880) rectangle (0.190, 0.920);
            \node[label={[label distance = 0cm]0:1}] at (0.180, 0.900) {};
            \draw[ultra thick,rounded corners] (0.001, 0.810) rectangle (0.223, 0.870);
            \node[label={[label distance = 0cm]:2}] at (0.112, 0.775) {};
            \draw[ultra thick,rounded corners] (0.001, 0.312) rectangle (0.223, 0.804);
            \node[label={[label distance = 0cm]:3}] at (0.112, 0.524) {};
            \draw[ultra thick,rounded corners] (0.001, 0.240) rectangle (0.223, 0.310);
            \node[label={[label distance = 0cm]0:4}] at (0.213, 0.265) {};
            \draw[ultra thick,rounded corners] (0.000, 0.000) rectangle (0.223, 0.092);
            \node[label={[label distance = 0cm]0:5}] at (0.213, 0.046) {};
            \draw[ultra thick,rounded corners] (0.223, 0.810) rectangle (0.303, 0.870);
            \node[label={[label distance = 0cm]:6}] at (0.313, 0.800) {};
            \draw[ultra thick,rounded corners] (0.224, 0.091) rectangle (1.000, 0.804);
            \node[label={[label distance = 0cm]0:7}] at (0.611, 0.543) {};
            \draw[ultra thick,rounded corners] (0.615, 0.000) rectangle (1.000, 0.092);
            \node[label={[label distance = 0cm]:8}] at (0.600, -0.010) {};
        \end{scope}
    \end{tikzpicture}

    \caption{The main GUI window, with enumerated sections} \label{fig: main gui with sections}

\end{mdframe}
\end{figure}
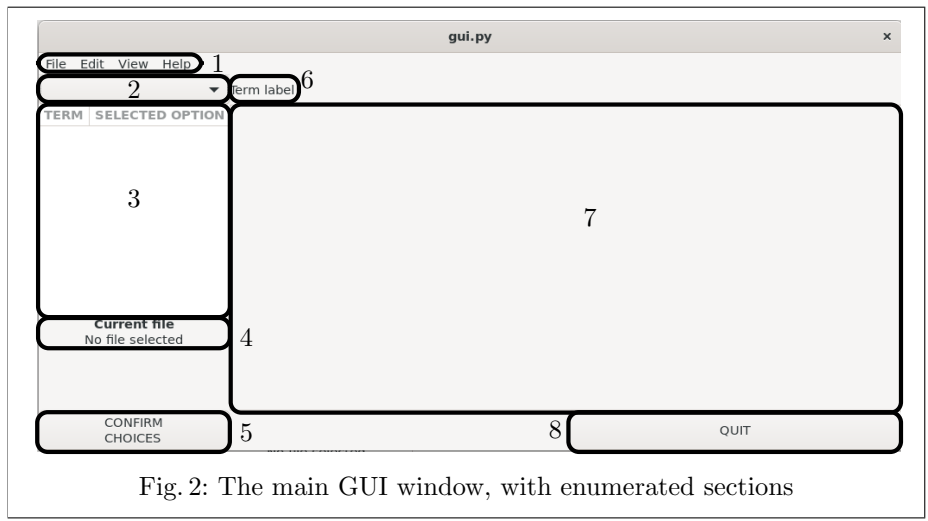

\def\figwidth{0.75\textwidth}
\begin{wrapfigure}[15]{r}{\figwidth}

\begin{mdframe}
    \begin{minipage}{\linewidth}
    
        \includegraphics[width=\linewidth]{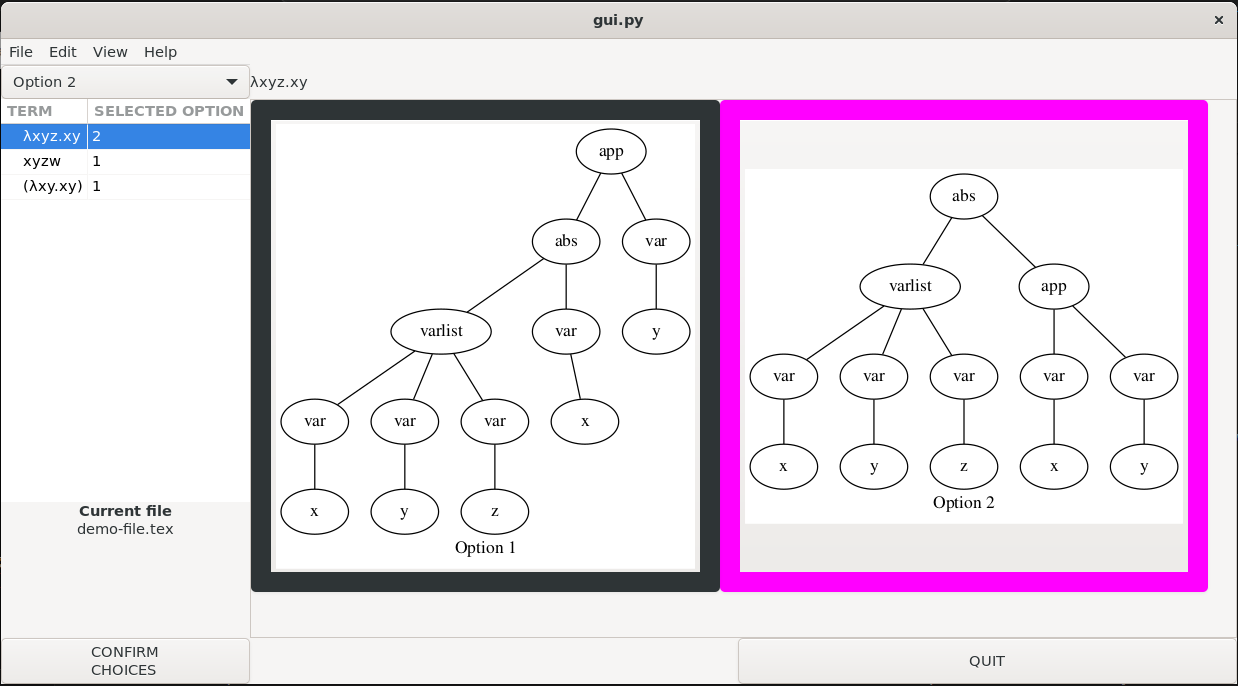}
        \caption{The GUI once a file has been opened} \label{fig: the gui with an open file}
    
    \end{minipage}
\end{mdframe}
\end{wrapfigure}

\Cref{fig: the gui with an open file} shows the GUI at work by parsing all the formulas found in \demofile\ using the grammar described in \Cref{example: grammar for lambda terms}.
We select the file we wish to \sTeX-ify by navigating to File$\to$Open in the toolbar.
The formula view shows 3 formulas that parsed ambiguously.
Once we select a formula, the ASTs for all the possible parses are shown in the grid layout.
The currently selected tree has a magenta outline, and the selection is also shown in the ``selected option'' column of the formula view.
To disambiguate a single formula, we need to either click the AST corresponding to the correct parse, or use the dropdown menu to select the correct option.
We then need to repeat this for all the formulas in the formula view, by clicking on each formula and then selecting the correct AST.

Once we select the correct parses for all formulas, we click the ``Confirm choices'' button, and the ASTs are converted into \sTeX\ macros (see \Cref{section: converting ASTs to stex macros}).
A new file called \texttt{demo-file-stex.tex} is produced, where each formula from \demofile\ is replaced with its \sTeX-ified counterpart.
\Cref{table: latex vs stex} shows all the formulas in \demofile\ and their corresponding \sTeX-ified versions.
\subsubsection{Visualising ASTs}
\setlength{\multicolsep}{0pt}
\setlength{\intextsep}{2pt}
\begin{wrapfigure}{R}{0.58\linewidth}

    \centering
    \begin{mdframe}
        \begin{minipage}{\linewidth}
        \begin{vwcol}[widths={0.65,0.35}, rule=0pt, sep=0pt, sidesep=0pt]
            \includegraphics[scale=0.57]{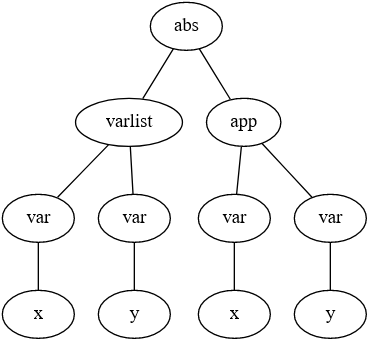}

            \begin{forest}
                indentation
                [abs
                [varlist
                [var
                [$x$]
                ]
                [var
                [$y$]
                ]
                ]
                [app
                [var
                [$x$]
                ]
                [var
                [$y$]
                ]
                ]
                ]
            \end{forest}
        \end{vwcol}
        \end{minipage}
        \captionsetup{skip=\baselineskip}
        \caption{Visualising the AST for $\protect\abs{\var{x}, \var{y}}{\app{\var{x}}{\var{y}}}$.} \label{fig: visualisation comparison}
    \end{mdframe}
    \end{wrapfigure}
To show the ASTs of all possible parses to users of the GUI, we need to visualize them into something easier to read.
We currently do this in two ways.
We can render them as SVGs using \texttt{graphviz} \cite{graphvizGraphviz2023}, by treating the nodes of the trees as nodes in a directed graph.
This allows them to be resized without loss of quality.

We can also use indentation to represent the hierarchy of a tree.
All the child nodes of a given parent have the same indentation.
For readability, we can also colour different ``levels'' in the tree with customizable colours (like some IDEs which apply different colours to nested parentheses, braces, etc.).
\Cref{fig: visualisation comparison} shows the different visualisations of the AST for the term $\abs{\var{x}, \var{y}}{\app{\var{x}}{\var{y}}}$.

\subsubsection{Converting ASTs to \sTeX\ Macros} \label{section: converting ASTs to stex macros}
\def\tablewidth{0.72\textwidth}
\begin{wraptable}[17]{r}{\tablewidth}
\vspace{-2ex}
\begin{mdframe}

    \caption{Original \LaTeX\ source code next to its \sTeX\ counterparts. A * indicates an ambiguous formula.} \label{table: latex vs stex}

    \begin{footnotesize}
        \begin{tabularx}{0.95\linewidth}{c|X}
            \hline
            \demofile & \texttt{demo-file-stex.tex} \\ \hline
            \mintinline[breaklines=true]{text}|*\lambda xyz.xy| & \mintinline[breaklines=true, breakafter={\{}]{text}|\abs{\var{x}, \var{y}, \var{z}}{\app{\var{x}}{\var{y}}}| \\ \hline
            \mintinline[breaklines=true]{text}|\lambda xy.x| & \mintinline[breaklines=true, breakafter={\{}]{text}|\abs{\var{x}, \var{y}}{\var{x}}| \\ \hline
            \mintinline[breaklines=true]{text}|y| & \mintinline[breaklines=true, breakafter={\{}]{text}|\var{y}| \\ \hline
            \mintinline[breaklines=true]{text}|xy| & \mintinline[breaklines=true, breakafter={\{}]{text}|\app{\var{x}}{\var{y}}| \\ \hline
            \mintinline[breaklines=true]{text}|*xyzw| & \mintinline[breaklines=true, breakafter={\{}]{text}|\app{\app{\app{\var{x}}{\var{y}}}{\var{z}}}{\var{w}}| \\ \hline
            \mintinline[breaklines=true]{text}|*(\lambda xy.xy)| & \mintinline[breaklines=true, breakafter={\{}]{text}|(\abs{\var{x}, \var{y}}{\app{\var{x}}{\var{y}}})| \\ \hline
            \mintinline[breaklines=true]{text}|\lambda xy.x| & \mintinline[breaklines=true, breakafter={\{}]{text}|\abs{\var{x}, \var{y}}{\var{x}}| \\ \hline
            \mintinline[breaklines=true]{text}|y| & \mintinline[breaklines=true, breakafter={\{}]{text}|\var{y}| \\ \hline
        \end{tabularx}
    \end{footnotesize}

\end{mdframe}
\end{wraptable}
Each AST can easily be converted into \sTeX\ macros.
The trees have been constructed with this in mind, so it is only a matter of ``flattening'' them into strings.
Special treatment (see \Cref{example: flexary argument with corresponding AST}) is needed for flexary arguments.

The \sTeX-ified formulas are inserted into a copy of the original document.
For each of them, the original formula is preserved in a comment right next to it in the source code.
If a formula could not be parsed, it is highlighted in red in the copied document.
Some ``boilerplate'' is also added, to enable \sTeX\ functionality.
This includes a \texttt{\footnotesize usepackage} command to include the \sTeX\ package, a \texttt{\footnotesize smodule} environment to wrap around the contents of the document\footnote{See Section 7.1 of the \sTeX\ manual \cite{kohlhaseSTEX3PackageCollection2023}}, and a \texttt{\footnotesize usemodule} command which imports the \lcalc\ macros we defined.
\Cref{table: latex vs stex} shows the formulas in \demofile\ next to their \sTeX-ified counterparts.

\subsubsection{Possible Improvements} \label{section: gui improvements}
In the future, we plan to add more options for displaying parse trees, such as displaying formulas with full parentheses (in this way, the nesting of parentheses reflects the structure of a given parse tree), and by ``joining'' the parse trees into forests similar to those produced by GLR parsers.
We will also investigate ways to handle large formulas which might not easily fit on the screen of the GUI, by possibly breaking them up into subformulas.
We can also improve the visual aspects of the GUI and the accessibility features, as currently everything is done by clicking with the mouse.
Support for customizable keyboard interactions will make using the GUI more user friendly, accessible, and potentially faster to use for those who prefer keyboard shortcuts.

\section{Grammar Generation}\label{section: grammar generation}
\def\u{\text{\textunderscore}}
\def\bs{\text{\textbackslash}}
To use the GUI to parse mathematical formulas, we need to supply a grammar to the parser.
Creating a single grammar to parse arbitrary formulas would be incredibly difficult and time consuming.
Even if we managed to create one, a large grammar is slower to parse with an exhaustive parsing algorithm like GLR, as the parser has to check a lot of possibilities that eventually get discarded.
As such, we believe that a modular approach is needed, to minimize wasted computation.

Even designing small grammars for specific portions of mathematical language is not easy.
It takes knowledge of grammar design, the relevant area of mathematics, and time to create such a grammar.
Designing grammars by hand would not scale well, so we started investigating automatic grammar generation.

The modularity of \sTeX\ archives and our intention to use \sTeX\ macros for semantic annotations inspired us to try generating grammars directly from \sTeX\ macro definitions.
The SMGloM \cite{kwarcSMGloMGitLab} already contains lots of definitions, so developing a way to systematically generate grammars from them could yield a highly modular and flexible tool for parsing and semantically annotating mathematical formulas.
This section describes the current state of our approach.

\subsection{Initial Approach}
\def\mintedsymdef{{\footnotesize\mintinline[breaklines=true]{text}|\symdef|}}
\def\mintedsymdecl{{\footnotesize\mintinline[breaklines=true]{text}|\symdecl|}}
\def\mintednotation{{\footnotesize\mintinline[breaklines=true]{text}|\notation|}}
First, let us consider an approach one might initially take to generating grammars from \sTeX\ macro definitions.
We could extract all the definitions provided by \mintedsymdef, \mintedsymdecl, and \mintednotation\ macros \footnote{See sections 7.2 and 7.4 of the \sTeX\ manual \cite{kohlhaseSTEX3PackageCollection2023}.} from a given \sTeX\ module.
Then, we could turn them into grammar rules by replacing all argument placeholders with a special nonterminal, \texttt{\footnotesize arg}, and turning all the \LaTeX\ code (which is just typesetting instructions) into terminals.

\def\plustexttt{\texttt{\footnotesize plus}}
\def\argtt{\texttt{\footnotesize arg}}
\def\arglambdatt{\texttt{\footnotesize arg\textunderscore setOfLambdas}}
\def\plustt{\texttt{\footnotesize "+"}}
If a semantic macro was defined as, e.g., {\footnotesize\mintinline[breaklines=true, breakbefore={\{}]{text}|\symdef{abs}[args=2]{\lambda #1 . #2}|}, a rule of the form {\footnotesize$\grule{\texttt{abs}}{\gseq{\texttt{"\textbackslash lambda"}, \argtt, \texttt{"."}, \argtt}}$} would be created.
Then, we can add a ``master rule'', with all the other nonterminals and a simple regular expression to match single letters\footnote{This is the last to be checked, only if none of the other nonterminals matched.} on the RHS: {\footnotesize$\grule{\argtt}{\galt{\ellipses, \texttt{abs}, \ellipses, \texttt{[a-z]+?}}}$}.
This approach quickly proves to be flawed.
It makes the assumption that anything is a valid argument to a semantic macro\footnote{In practice, this is not the case. For example, a semantic macro for addition of natural numbers should not take sets as arguments.}, which leads to \textit{over-generating} grammars and a large number of nonsensical trees.
It is evident that the approach has to be refined.
\begin{mdframe}[innertopmargin=3pt]
\begin{minipage}{\linewidth}
    \begin{sexample}[title={An example grammar}]
        
        \begin{minipage}{\linewidth}

        \begin{wrapfigure}[9]{r}{0.45\linewidth}
        
            \begin{ttmath}
                \begin{footnotesize}
                    \setlength{\abovedisplayskip}{0pt}
                    \setlength{\belowdisplayskip}{0pt}
                    \begin{align*}
                        & \grule{arg}{\galt{var, abs, app, text}} \\
                        & \grule{var}{arg} \\
                        & \grule{abs}{\gseq{``\bs lambda", arg, ``.", arg}} \\
                        & \grule{app}{\gseq{arg, arg}} \\
                        & \grule{text}{[a-z]+?} \\
                    \end{align*}
                \end{footnotesize}
            \end{ttmath}
            \captionsetup{skip=-1.5\baselineskip}
            \caption{A grammar generated from \\the macros for the \lcalc}
            \label{fig: grammar from lcalc macros}
        \end{wrapfigure}

        Recall the semantic macros defined in \Cref{section: stex macros for the lcalc}.
        For this example, we simplified the abstraction macro by replacing the flexary argument with a regular one.
        Our initial approach generates the grammar in \Cref{fig: grammar from lcalc macros}.
        It is clear that the grammar will over-generate, as an abstraction or application can appear as the first argument of an abstraction (which should only be a variable).
    \end{minipage}
    \end{sexample}
\end{minipage}
\end{mdframe}

\subsection{Adding Types}\label{section: adding types to grammar}
An option to provide a \textit{type} to a macro is available in \sTeX.
We can extract types from macro definitions and use them to restrict the possible arguments of a macro.
\begin{mdframe}[innermargin=0pt]
\begin{minipage}{\linewidth}
    \begin{sexample}[title={Macro with a type and the corresponding grammar rules}]

        Suppose that we have a macro definition of the form {\footnotesize\mintinline[breaklines=true]{text}|\symdef{app}[name=application, args=2, type=\funspace{\setOfLambdas, \setOfLambdas}{\setOfLambdas}]{#1 #2}|}.
        Its type is $\funspace[arrowtimes]{\setOfLambdas, \setOfLambdas}{\setOfLambdas}$, it takes two \lambdaterms\ as input and returns a \lambdaterm.
        We can use this information to restrict the arguments in the grammar rule corresponding to {\footnotesize\mintinline[breaklines=true]{text}|\app|}.
        The rules we generate from it are {\footnotesize$\grule{\texttt{app}}{\gseq{\arglambdatt, \plustt, \arglambdatt}}$} and {\footnotesize$\grule{\arglambdatt}{\texttt{app}}$}.
    \end{sexample}
\end{minipage}
\end{mdframe}

For each type present in the grammar, we also need to provide a default to match if all else fails (like the letter-matching regular expression in \Cref{fig: grammar from lcalc macros}).
We can use \parglare's \textit{recognizers} for this, but they would have to be manually specified.
For the \arglambdatt\ rule, the recognizer can be a simple function that finds uppercase letters possibly followed by apostrophes or subscripted indexes.

There are two issues to solve before we can scale this to more areas of mathematics.
First, not all macro definitions in the SMGloM provide a type signature.
Adding them for all macros is a difficult task without automation of type checking, which requires a particular choice of type system to serve as a foundation.
Additionally, it is not always obvious what the type of a mathematical object (and thus the semantic macro representing it) should be.

\subsubsection{Types in a More ``Loose'' Sense}
We could add types in a less strict sense, akin to some programming languages.
When defining a function in, e.g., Java, we need to specify the types for all its arguments, and the type of any value that is returned.
We could take a similar approach to typing when generating grammars, by specifying what a valid argument for each macro is.
A grammar could be generated using any available types.
Then, for rules which have no type restrictions for their arguments, we can prompt the user to select all the nonterminals or recognizers of the correct ``type''.
\begin{mdframe}
\begin{sexample}[title={Adding types to an ``untyped'' rule}]
    Recall the grammar rule for abstractions from the grammar in \Cref{fig: grammar from lcalc macros}.
    We can restrict its first argument to only accept variables by changing the rule for \texttt{abs} to\linebreak{\footnotesize$\grule{\texttt{abs}}{\gseq{\texttt{"\bs lambda"}, \texttt{abs\u arg\u 1}, \texttt{"."}, \texttt{abs\u arg\u 2}}}$}, and adding the rules \linebreak {\footnotesize$\grule{\texttt{abs\u arg\u 1}}{\texttt{var}}$} and {\footnotesize$\grule{\texttt{abs\u arg\u 2}}{\texttt{arg}}$}\footnotemark.
\end{sexample}
\end{mdframe}
\footnotetext{The second rule is purely illustrative. In practice, we would restrict this grammar further by e.g., adding a recognizer for variables and possibly specifying a general nonterminal (and semantic macro) for a general \lambdaterm\ (like the \texttt{lexp} in the grammar shown in \Cref{example: grammar for lambda terms}).}
\subsection{Using Precedence to Restrict Grammars Further}
We can use \textit{precedence} to restrict grammars even further.
If rules have different priorities assigned, \parglare\ uses them to disambiguate parses by considering only the rule with the highest priority.
In \sTeX, precedence provides information about how strong a macro binds its arguments, which is used e.g., for automated bracketing.
Lower numbers\footnote{Precedences in \sTeX\ can be negative and range from $-2^{32}$ to $2^{32}$.} indicate stronger binding and thus higher precedence.
We can use \sTeX's precedence to assign priority to grammar rules, but we must first map them to non-negative integers used by \parglare.

\subsection{Standardising Grammars and Parse Actions}
\def\notationnonterm#1#2{{\footnotesize\texttt{#1\textunderscore#2\textunderscore rule}}}
\def\mainmacrononterm#1{{\footnotesize\texttt{main\textunderscore#1\textunderscore rule}}}
Besides improving the initial approach to grammar generation using types and precedence, we also standardised parse actions and the naming of nonterminals to follow a specific pattern.
Each semantic macro definition (via \mintedsymdef\ or \mintedsymdecl) provides the name of a macro, e.g. {\footnotesize\texttt{macroname}} (abbreviated as \texttt{\footnotesize mn}).
We use that to create a nonterminal with the name \mainmacrononterm{macroname}.
For each notation for a given \texttt{\footnotesize mn} (defined via \mintedsymdef\ or \mintednotation), we create a \textit{notation rule} of the form $\grule{\notationnonterm{mn}{name}}{\texttt{\footnotesize RHS}}$, where \texttt{\footnotesize name} is the name of the notation (this is an empty string if the notation is defined using \mintedsymdef).
We generate \texttt{\footnotesize RHS} by replacing argument placeholders with nonterminals representing types, as per \Cref{section: adding types to grammar}.
Then, a rule of the form $\grule{\mainmacrononterm{mn}}{\galt{\notationnonterm{mn}{n1}, \ellipses, \notationnonterm{mn}{nN}}}$ is created, where each \notationnonterm{mn}{nK} is the LHS of a notation rule.
\begin{mdframe}
\begin{sexample}[title={Automatic grammar gemeration from a sample archive}]\label{example: generated grammar}
    After including precedences and types, our macros for annotating \lambdaterms\ from \Cref{table: semantic macro definitions} now look like this:
\begin{footnotesize}
\begin{minted}[breaklines=true, breakafter={\],}, breakbefore={\{}]{text}
\symdef{var}[name=variable, args=1, type=\varSet]{#1}
\symdef{abs}[name=abstraction, args=ai, prec=51;\infprec x\infprec, type=\funspace{\varSet, \setOfLambdas}{\setOfLambdas}]{\maincomp{\lambda}\argsep{#1}{}\comp{.}#2}
\symdef{app}[name=application, args=2, prec=50;50x49, type=\funspace{\setOfLambdas, \setOfLambdas}{\setOfLambdas}]{#1 #2}
\end{minted}
\end{footnotesize}
    Here, {\footnotesize\verb|\varSet|} and {\footnotesize\verb|\setOfLambdas|} are used to represent $\varSet$ and $\setOfLambdas$, respectively.
    In this way, we limit the first argument of an abstraction to only be a variable, while the second argument can be any \lambdaterm.
    Similarly, both arguments of an application are now necessarily \lambdaterms.
    We can then automatically generate a grammar (in \parglare\ syntax) from these macro definitions.
    To use the grammar to parse \demofile\ with \parglare, some manual tweaks are required\footnotemark.
    Removals from the grammar are indicated in red, while additions are indicated in green.
    Numbers in braces indicate precedence, and expressions of the form \texttt{foo=bar} allow us to pass the value of \texttt{bar} to parse actions as a \textit{keyword argument} with name \texttt{foo}.
\begin{footnotesize}
\begin{minted}[breaklines=true, escapeinside=!!]{text}
main_grammar: main_var_rule {10} | main_abs_rule {11} | main_app_rule {12} | !\textcolor{red}{text |}! par_exp;
var__arg_1: varSet_type_rule;
varSet_type_rule: !\textcolor{red}{var{\_\_}rule |}! varSet_type_rule_recognizer !\textcolor{red}{| variable{\_}recognizer}!;
var__rule:  arg1=var__arg_1 ;
main_var_rule: var__rule {10};
abs__arglist_1: abs__arg_1  abs__arglist_1 | abs__arg_1;
abs__arg_1: !\textcolor{red}{setOfLambdas{\_}type{\_}rule} \textcolor{Green}{varSet{\_}type{\_}rule}!;
!\textcolor{red}{setOfLambdas\_type\_rule: abs{\_\_}rule | app{\_\_}rule | setOfLambdas{\_}type{\_}rule{\_}recognizer | variable{\_}recognizer;}!
!\textcolor{Green}{setOfLambdas\_type\_rule: main\_var\_rule \{10\} | main\_abs\_rule \{11\} | main\_app\_rule \{12\} | setOfLambdas\_type\_rule\_recognizer;}!
abs__arg_2: !\textcolor{red}{main\_grammar} \textcolor{Green}{setOfLamdas\_type\_rule}!;
abs__rule:  bs "lambda" arg1=abs__arglist_1  dot  arg2=abs__arg_2 ;
main_abs_rule: abs__rule {11};
app__arg_1: !\textcolor{red}{main\_grammar} \textcolor{Green}{setOfLamdas\_type\_rule}!;
app__arg_2: !\textcolor{red}{main\_grammar} \textcolor{Green}{setOfLamdas\_type\_rule}!;
app__rule:  arg1=app__arg_1  arg2=app__arg_2 ;
main_app_rule: app__rule {12};
par_exp: "(" main_grammar ")";
any_type_rule: main_grammar !\textcolor{red}{| variable\_recognizer}!;

terminals
bs: "\\";
dot: ".";
text: /[A-z]+?/;
varSet_type_rule_recognizer: !\textcolor{Green}{/[a-z]+?/}!;
setOfLambdas_type_rule_recognizer: !\textcolor{Green}{/[A-Z]+?/}!;
!\textcolor{red}{variable\_recognizer: ;}!    
\end{minted}    
\end{footnotesize}
    Some rules (such as the rules for \texttt{\footnotesize par\u exp} and \texttt{\footnotesize dot}) are hardcoded into this grammar, as they would be needed in most grammars we generate.
    For brevity, we use regexes instead of recognizers for variables and general \lambdaterms, since variables in \demofile\ do not contain any apostrophes or subscripts.
    This grammar can now be used to parse \demofile, and ambiguities can be resolved using the GUI.
    The ambiguities that occur are the same as if we used the manually specified grammar from \Cref{example: grammar for lambda terms}.
    The formulas and their \sTeX-ified counterparts are laid out in \Cref{table: latex vs stex}.
\end{sexample}
\end{mdframe}
\footnotetext{Most of these can be made obsolete by improving the grammar generation code, in particular when it comes to using types. This will be addressed in future work.}
\subsubsection{Standardising Parse Pctions}
By standardising the grammar structure and naming of nonterminals, we only need a handful of parse actions for transforming the parse results into ASTs, which we assign to certain kinds of nonterminals.

Nonterminals of the form {\footnotesize\mainmacrononterm{mn}} do not contribute to the structure of the AST, so we can define a parse action that returns the subtree without changing it.
Nonterminals of the form {\footnotesize\texttt{mnn\u name}} ({\footnotesize\texttt{mnn}} is used as a shorthand for {\footnotesize\texttt{macroname\u notation\u name}}) return an AST with a node named {\footnotesize\texttt{mnn\u rule}} as the root, and the arguments of that macro as children.

\begin{mdframe}
\begin{minipage}{\linewidth}
    \begin{sexample}[title={AST returned by matching a notation}]

        \begin{minipage}{\linewidth}
            If we wish to annotate {\footnotesize\mintinline[breaklines=true]{text}|\lambda x.x|}, a portion of the parse tree is given by matching the rule {$\grule{\footnotesize \texttt{abs\u\u rule}}{\footnotesize \gseq{\texttt{bs "lambda"\!}, \texttt{abs\u\u arglist\u 1\!}, \texttt{dot\!}, \texttt{abs\u\u arg\u 2}}}$}.
            \begin{wrapfigure}{r}{0.30\textwidth}
                \centering
                \begin{forest}
                    for tree={l=4pt}
                    [\texttt{\footnotesize abs}
                    [$x$]
                    [$x$]
                    ]
                \end{forest}
                \caption{Example AST} \label{fig: parse action example}
            \end{wrapfigure}
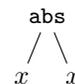
            The resulting AST (see \Cref{fig: parse action example}) should not include {\footnotesize\texttt{bs "lambda"}} or {\footnotesize\texttt{dot}}, as they are irrelevant to the overall structure of the term.
            We can use a parse action to build the AST, with the LHS of the rule as the root, and the nonterminals on the RHS that represent arguments, as children.
            This closely resembles the structure of the semantic macro that we should use to annotate the input string, {\footnotesize \mintinline{text}|\abs{x}{x}|}.
        \end{minipage}
    \end{sexample}
\end{minipage}
\end{mdframe}
\def\arglistnonterm{\texttt{mnn\u arglist\u N}}
\def\arglistelnonterm{\texttt{mnn\u arg\u N}}
Special treatment is required for flexary arguments.
Grammar rules of the form ${\footnotesize \grule{\arglistnonterm}{\galt{\gseq{\arglistelnonterm, \texttt{sep}, \arglistnonterm}, \arglistelnonterm}}}$ are used for flexary arguments, and produce distinctly shaped parse trees.
We transform them into an AST with a single parent node and one ``layer'' of children (see \Cref{example: flexary argument with corresponding AST}).

\def\vartt{\texttt{var}}
\def\varlisttt{\texttt{varlist}}
\begin{mdframe}
    \begin{minipage}{\linewidth}
    \begin{sexample}[title={A parsed flexary argument and the corresponding AST}] \label{example: flexary argument with corresponding AST}
        
        \begin{minipage}{\linewidth}
            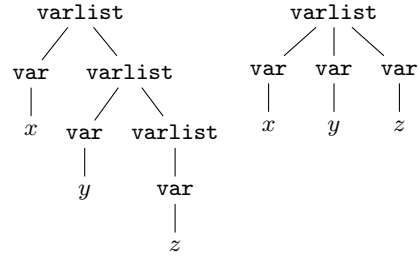
\begin{wrapfigure}{r}{0.52\linewidth}
                \begin{multicols}{2}
                    \footnotesize
                    \begin{forest}
                        for tree={l=4pt, s sep=4pt}
                        [\varlisttt
                        [\vartt
                        [$\var{x}$]]
                        [\varlisttt
                        [\vartt
                        [$\var{y}$]]
                        [\varlisttt
                        [\vartt
                        [$\var{z}$]
                        ]
                        ]
                        ]
                        ]
                    \end{forest}
                    
                    \begin{forest}
                        for tree={l=4pt, s sep=4pt}
                        [\varlisttt
                        [\vartt
                        [$\var{x}$]]
                        [\vartt
                        [$\var{y}$]]
                        [\vartt
                        [$\var{z}$]]
                        ]
                    \end{forest}
                \end{multicols}
                \caption{The parse tree (left) and the AST we wish to obtain from it (right)}
                \label{fig: parse tree and AST of flexary argument}
    \end{wrapfigure}
    
    Recall the \mintinline{text}|\abs| macro defined in \Cref{table: semantic macro definitions}.
    Its first argument is a flexary sequence of variables that can be parsed by a grammar rule of the form ${\footnotesize \grule{\varlisttt}{\galt{\gseq{\vartt, \varlisttt}, \vartt}}}$ (see \Cref{example: grammar for lambda terms}).
    For brevity we are omitting the rules for parsing \vartt.
    When parsing terms like $\abs{\var{x}, \var{y}, \var{z}}{A}$, we obtain very ``right-leaning'' parse trees, which we have to transform into ASTs with one ``layer'' of children (see \Cref{fig: parse tree and AST of flexary argument}).
\end{minipage}
\end{sexample}
\end{minipage}
\end{mdframe}

Standardising nonterminal names and actions yields ASTs which can easily be converted to \sTeX\ macros.
The names of nonterminals in an AST contain the names of relevant semantic macros and notations.
Standardising nonterminal names will also ensure fewer conflicts when we merge multiple grammars to parse a document.
The only issue to solve is two \sTeX\ archives defining a macro with the same name.
This can be avoided by attaching metadata to nonterminals\footnote{This is another feature of \parglare.}.

\subsection{Current State and Issues to Address} \label{section: current state and issues}
The grammars we produce can be used to parse small formulas and produce semantically annotated equivalents.
We have only done limited testing, because the generated grammars can contain \textit{left recursion} or \textit{cycles}, e.g., in \Cref{example: generated grammar},
{\footnotesize$\grule{\texttt{var\u\u rule}}{\grule{\texttt{var\u\u arg\u 1}}{\grule{\texttt{varSet\u type\u rule}}{\texttt{var\u\u rule}}}}$}.
For clarity, we only show the nonterminals that form this cycle.
In the example, we had to manually modify the grammar to remove such cycles.
We can potentially eliminate some cycles by improving the grammar generation code, especially when it comes to providing rules for a given type.
In \Cref{example: generated grammar}, the rule for \texttt{\footnotesize setOfLambdas\_type\_rule} did not have the ``main'' nonterminals for a given macro on the RHS, but rather nonterminals associated with a specific notation for a macro.

However, eliminating all cycles during grammar generation is not a guarantee\footnote{Even after modifying the grammar in \Cref{example: generated grammar}, we still have the cycle {\footnotesize$\grule{\texttt{abs\u\u arg\u 2}}{\grule{\texttt{setOfLambdas\u type\u rule}}{\grule{\texttt{main\u abs\u rule}}{\grule{\texttt{abs\u\u rule}}{\texttt{abs\u\u arg\u 2}}}}}$}, for example.} and \parglare\ cannot handle cyclic grammars.
This leaves us with two options: (1) modify the grammars to remove cycles, or (2) use a parser that handles cyclic grammars.
Algorithms for removing cycles exist, but they change the structure (and size) of the grammar.
We wish to avoid that, as ``cyclic grammars can be the most compact way to describe a language'' \cite{rekersParserGenerationInteractive1992}.
Thus we will address this issue in the parser itself.
This will require extensive research of the state of the art, or adapting existing parsing algorithms.

\subsubsection{Adding Types to Grammars}
Not many semantic macro definitions have types added to them.
As such, we might have to consider manually adding ``types'' of some sort to grammars that we generate from ``untyped'' semantic macro definitions.
This could be as simple as just editing the grammar files, or we could develop a special interface for the task.
Each grammar we create could then be shared in an open-source repository similar to the SMGloM, to avoid a reduplication of effort from other users trying to \sTeX-ify similar documents.

\subsection{Alternatives}
We could search for existing grammars and manually adapt them to fit with our work, and connect them with \sTeX\ archives based on whether the parse results can be represented with macros from a given archive.
We could, however, lose the one-to-one correspondence between grammars and \sTeX\ achives that we have by generating grammars using the approach we presented in this paper.

We could also ``crowdsource'' the grammars by starting a community-developed collection of small grammars based on \sTeX\ archives.
This would require precise instructions and rigorous checks for contributors, to avoid e.g., name conflicts in nonterminals, and ensure the grammars can seamlessly be used simultaneously.

\section{Conclusion and Future Work} \label{section: conclusion}

In this paper we presented the exploration of a novel approach to \sTeX-ification by generating grammars from \sTeX\ macro definitions to parse a document and using a GUI to disambiguate parses.
We have shown how it can help \sTeX-ify mathematical documents on a practical example.
We also discuss a method to generate grammars for use with the GUI, and elaborated on the issues we face.

In the future, we plan on doing at least the following: 
(1) as mentioned in \Cref{section: gui improvements}, we will add more features to the GUI;
(2) as mentioned in \Cref{section: current state and issues}, we will address the issue of cyclic grammars, and develop tools to speed up the grammar creation process;
(3) we will develop an interface with which users can select the macros/archives they wish to use in their document, from which a grammar is to be generated
(4) we will look into ways to share grammars between users, to avoid the reduplication of effort;
(5) we will develop a process for merging the smaller grammars we generate into grammars that can handle larger, less specific documents than \demofile;
(6) we will generalise the ``boilerplate'' added to an \sTeX-ified document to add all modules whose macros are being used;
(7) we will add support for author-defined macros (this might take the form of a preprocessing step that selectively expands them);
(8) we will investigate ways to (semi-)automatically annotate other parts of a document, e.g., definitions, examples, and textual references to mathematical objects.

\bibliography{references}

\end{smodule}
\end{document}